# Human Detection and Tracking for Video Surveillance: A Cognitive Science Approach

Vandit Gajjar
L. D. College of Engineering
Ahmedabad
`gajjar.vandit.381@ldce.ac.in`

Ayesha Gurnani
L. D. College of Engineering
Ahmedabad
`gurnani.ayesha.52@ldce.ac.in`

Yash Khandhediya
L. D. College of Engineering
Ahmedabad
`khandhediya.yash.364@ldce.ac.in`

**Abstract**

*With crimes on the rise all around the world, video surveillance is becoming more important day by day. Due to the lack of human resources to monitor this increasing number of cameras manually, new computer vision algorithms to perform lower and higher level tasks are being developed. We have developed a new method incorporating the most acclaimed Histograms of Oriented Gradients, the theory of Visual Saliency and the saliency prediction model Deep Multi-Level Network to detect human beings in video sequences. Furthermore, we implemented the k – Means algorithm to cluster the HOG feature vectors of the positively detected windows and determined the path followed by a person in the video. We achieved a detection precision of 83.11% and a recall of 41.27% in a time as short as 23.0047 seconds.*

## 1. Introduction

Computer Vision is day by day becoming important and with that human detection for applications like video surveillance, autonomous driving vehicles, person recognition have also become important. Human Detection is challenging because everyone is different in appearances and there are wide range of poses. There should be a robust method for feature extraction even when the background is cluttered. The cameras used for these applications make use of RGB cameras during night when there is deficiency of light and the images are not clear. This makes the changes in lighting conditions an important point as well. For this, many researchers have proposed different methods for detecting humans from any image. Different methods for extraction of features have been developed so that the features can be applied to SVM for classification. One of the proposed methods is Histogram of Oriented Gradients [2], which is extensively used in detection of human beings.

Visual Saliency is computed using various different methods after years of research on the field of cognitive science. It is still an open problem with the MIT Saliency Benchmark [7] getting newer and accurate models as time passes. Visual Saliency is basically an intensity map where higher intensities signify regions where a general human being would look and lower intensities mean decreasing level of visual attention. It is a measure of visual attention of humans on the basis of the content of the image. Here, we have used visual saliency for the purpose of region proposal. We used the Deep Multi-Level Network [1] developed for the same purpose. It has a encoder-decoder architecture which computes saliency maps as outputs.

We used the HOG features to train a Support Vector Machine classifier [11] for detecting human beings in any frame. For this we used the OSU Color-Thermal dataset from the Ohio State University [4]. Further, we used the salience-windowed images for classification after training. With this, we got reduced computational time. This makes the process lot faster. But this isn't enough. The motion pattern of the human beings in the frame is equally important and we have addressed that problem as well.

The k - Means algorithm [5] is an algorithm used to determine clusters of various data-points in the given data. Clustering can come in handy when we need to put similar things together in a same class. Data-points which are closer to each other on the data-space are considered to be in the same cluster. Using the k - means algorithm enabled us to find the movement patterns of the humans in the frame. This is important for person re-identification which is in itself another active research problem. We used the k – means algorithm to compare and cluster HOG features from consecutive frames so as to identify the set of points on the image resembling a particular person moving in the video.

## 2. Related Work

Various methods have been used time and again for human detection and tracking in videos. These methods include the most acclaimed Histogram of Oriented Gradients as feature extractor [2]. Other than that, there are many different works which developed novel approaches within the scope of HOG [3, 16]. Few other methods have also been developed which use probabilistic assembly of robust part detectors [8], classification on riemannian manifolds [12], depth information [13], flexible mixture of parts [14], etc.

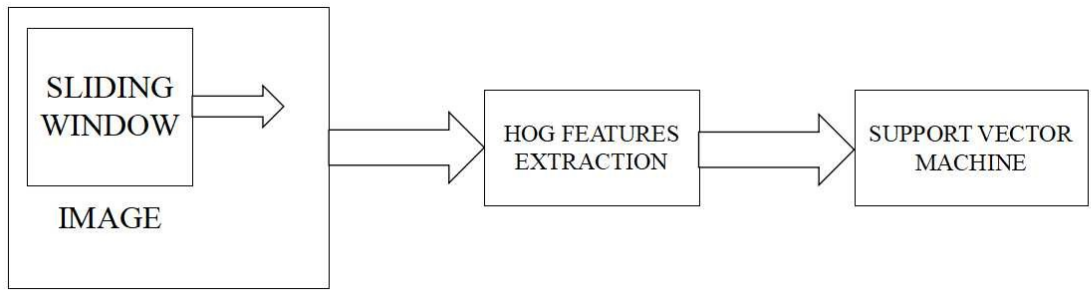

Figure 1: Block diagram for Human Detection in Normal Images using HOG Features

Newer and better deep learning models have also been developed [10, 15]. The scope of this paper is limited to developing further on the HOG [2].

## 3. The Dataset

We used the OSU Color-Thermal Pedestrian Dataset [4] from the Ohio State University. It is a dataset collected on campus of the university. It includes mainly the campus interaction of people from a birds-eye view which is basically what we want. All video surveillance cameras are placed in such a way that no person can reach to them. Therefore, they are kept at a larger height. The images are of size 320x240 pixels amounting to a total of 17089 images. It was collected at three different locations on the campus of OSU thereby including variation in scene as well. Each image was captured periodically with a time period of 1/30 seconds. The dataset also includes Thermal images but we have only used the RGB images for this research.

## 4. Human Detection using HOG features

We firstly emulated the results achieved by [2] for comparison purpose. We trained the Support Vector Machine (SVM) classifier [11] as a two class classifier on the dataset by giving cropped human images as positive samples and random images of the same size as negative samples. The classification accuracy on the training dataset was found to be very high in the order of 98.32%.

We tested the SVM classifier on other unseen images of the dataset. To detect humans from the whole image, the basic algorithm assumes a window of a particular size. Windows of this size are cropped out from the image one by one in a convolutional manner and HOG features are computed as shown in Figure 1. As the size of the image increases, total number of windows to be classified increases and due to extraneous data, the computational time to compute HOG features as well as for classification increases as well. The computation time taken was 1768.29 seconds for a single image in the best case. This method is hence not viable for real-time applications.

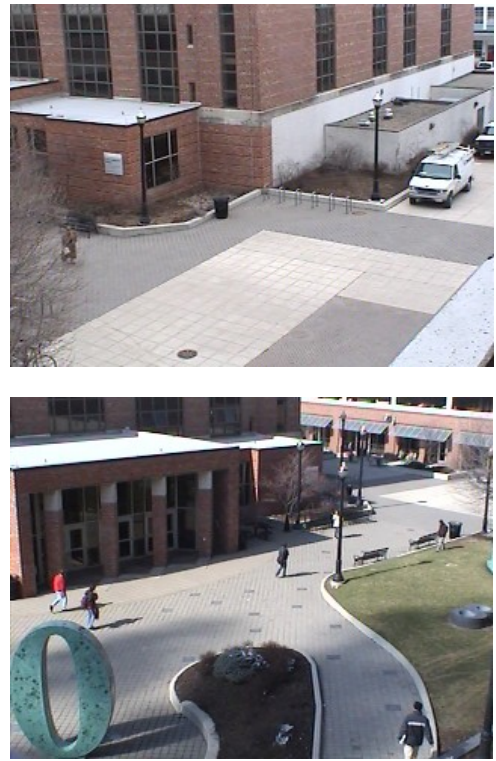

Figure 2: Sample Images from the Dataset

Regardless, we evaluated the results of the classification and found that the many false positive existed in the inferences of the classifier. This was due to vaguely similar HOG features of some parts in the background with the HOG features of a general human being. The recall was computed to be 0.93 which is very high. A proper model for human detection in surveillance videos required a perfect blend of precision as well as recall. Therefore, we did a bit of research to find new region proposal algorithms which would be helpful in reducing the recall as well as the computation time.

## 5. The Saliency Model

We used the Deep Multi-Layer Network for saliency prediction [1]. It is called the ML – Net. The purpose of using a Visual Saliency model was to propose possible regions where humans might be in the frame. Visual Saliency maps would indicate higher intensity where there are humans in the image because of results supported by [6]. An example of the salience-windowed image is given in Figure 3.

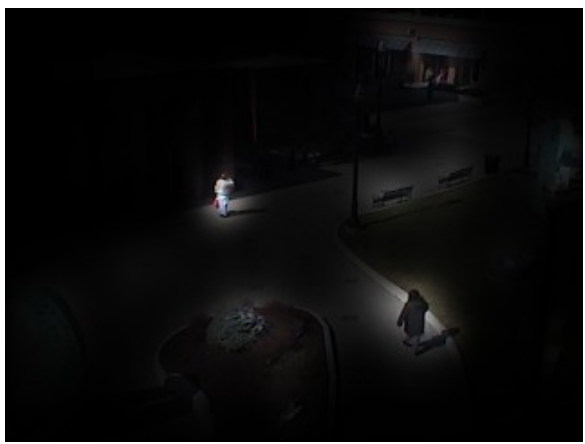

Figure 3: Example of Salience-windowed image

The Deep Multi-Layer Network described in [1] outperforms all other models on the SALICON Dataset and also performs better on the MIT Saliency Benchmark [7]. This model is sufficient for our use i.e. human detection for video surveillance.

## 6. Experiments

We did various different experiments using the visual saliency maps for region proposal [9]. Firstly, we tried to make a bounding box out of salient regions in the saliency maps. This method was not successful because of the fading out of salient regions through the borders.

We ended up multiplying the saliency maps with the images making the images salience-windowed. An example of the salience-windowed image is given in Figure 3. This gave us the reduced data that we were looking for to reduce the computation time as well as the recall. The comparison of results of classification on simple images and salience-windowed images is given in Table 1.

| Parameters | Normal Images | Salience-windowed Images |
|---|---|---|
| Execution Time (in seconds) | 1768.29 | 23.0047 |
| Precision | 12.36% | 83.11% |
| Recall | 93.12% | 41.27% |

Table 1: Comparison of Results

Excellent performance was observed on the salience-windowed images as compared to the previous experimentation on normal images. Visual Saliency as a region proposal algorithm works great for human detection using HOG features. The complete block diagram of the method can be seen in Figure 4.

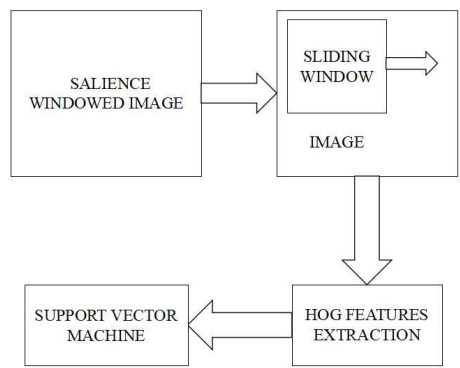

Figure 4: Block Diagram of Salience-windowed Image

## 7. Post Recording Tracking

Tracking of human beings is important as well when it comes to detecting humans. Addressing this problem solves many woes of the security team. We have used the k – Means Clustering algorithm [5] to track particular humans in the frame and also get to know their walking path.

We first record the location of each person in the frame for every successive frame. This is basically recording the location of every detected person on the frame. With this, we also recorded their particular HOG feature vectors associated with their location. Due to this, we call this method the post recording tracking.

The maximum number of detected persons in any frame among the recorded data becomes the value of 'k' for tracking. We randomly initialize the algorithm 100 times and consider only the one which offers the least distance from the mean points of the recorded data.

Each cluster of HOG features associate with location on frame represent the set of points on the path of a single person. We can then easily connect the points of each cluster thus giving us the path followed by each person in the frame. The block diagram of the method can be seen in Figure 5.

Motion tracking hereby becomes possible due to implementation of the k – Means algorithm to determine clusters of feature points representing the path followed by a person.

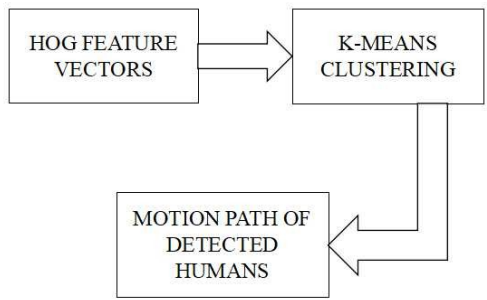

Figure 5: Block Diagram of Post Recording Tracking

## 8. Conclusion

We observed that detection and motion patterns of humans can be found out efficiently by using the proposed method in this paper. Using Visual Saliency as a region proposal algorithm proved to be beneficial for the research work. Human Detection improved by introducing the salience-windowed frames of the video to the HOG + SVM classifier [2]. The performance improved manifolds as compared to classification on normal images. This happened due to unnecessary data which was in a way increasing the recall of the model and also the computation time and thus making the model not fit to be used in real time. Using the Deep Multi-Layer Network for saliency prediction [1] to propose regions of interests makes the model much more feasible in terms of computational requirements as well as precision and recall.

We also observed the usage of k – Means algorithm in clustering feature vectors of the same person and how it can be used to compute the path followed by a person in the video sequence.

## 9. Future Work

We are planning to incorporate motion tracking method Optical Flow in conjunction with Visual Saliency windowing to detect motion patterns of human beings for video surveillance.

## 10. Acknowledgments

We are thankful to the anonymous reviewers for their valuable comments due to which the paper was improved. We are also thankful to Prof. Usha Neelakantan, Head of Department, Electronics and Communication Engineering, Lalbhai Dalpatbhai College of Engineering, Ahmedabad for her relentless support throughout the period of the project.